\documentclass{article}

\usepackage{microtype}

\usepackage{float}
\usepackage{graphicx}
\usepackage{subfigure}
\usepackage{booktabs} 

\usepackage{hyperref}

\usepackage[accepted]{icml2025}

\usepackage{amsmath}
\usepackage{amssymb}
\usepackage{mathtools}
\usepackage{amsthm}

\usepackage[capitalize,noabbrev]{cleveref}

\theoremstyle{plain}
\newtheorem{theorem}{Theorem}[section]
\newtheorem{proposition}[theorem]{Proposition}

\theoremstyle{definition}

\theoremstyle{remark}

\usepackage[textsize=tiny]{todonotes}

\icmltitlerunning{Train Once, Forget Precisely: Anchored Optimization for Efficient Post-Hoc Unlearning}

\begin{document}

\twocolumn[
\icmltitle{Train Once, Forget Precisely: Anchored Optimization for Efficient Post-Hoc Unlearning}



\icmlsetsymbol{equal}{*}

\begin{icmlauthorlist}

\icmlauthor{Prabhav Sanga}{equal,ucl,thinkai}
\icmlauthor{Jaskaran Singh}{equal,uon,thinkai}
\icmlauthor{Arun Kumar Dubey}{bvce,thinkai}

\end{icmlauthorlist}

\icmlaffiliation{ucl}{University College London, London, United Kingdom}
\icmlaffiliation{uon}{University of Nottingham, Nottingham, United Kingdom}
\icmlaffiliation{thinkai}{Think-AI, New Delhi, India}
\icmlaffiliation{bvce}{Bharati Vidyapeeth College of Engineering, New Delhi, India}

\icmlcorrespondingauthor{Arun Kumar Dubey}{arun.dubey@bharatividyapeeth.edu}

\icmlkeywords{Machine Learning, ICML}

\vskip 0.3in
]



\printAffiliationsAndNotice{}  

\begin{abstract}
As machine learning systems increasingly rely on data subject to privacy regulation, selectively unlearning specific information from trained models has become essential. In image classification, this involves removing the influence of particular training samples, semantic classes, or visual styles without full retraining. We introduce \textbf{Forget-Aligned Model Reconstruction (FAMR)}, a theoretically grounded and computationally efficient framework for post-hoc unlearning in deep image classifiers. FAMR frames forgetting as a constrained optimization problem that minimizes a uniform-prediction loss on the forget set while anchoring model parameters to their original values via an $\ell_2$ penalty. A theoretical analysis links FAMR's solution to influence-function-based retraining approximations, with bounds on parameter and output deviation. Empirical results on class forgetting tasks using CIFAR-10 and ImageNet-100 demonstrate FAMR's effectiveness, with strong performance retention and minimal computational overhead. The framework generalizes naturally to concept and style erasure, offering a scalable and certifiable route to efficient post-hoc forgetting in vision models.
\end{abstract}

\section{Introduction}

As machine learning systems become increasingly pervasive in sensitive domains, such as medical diagnostics and user-facing recommendation engines, ensuring compliance with privacy regulations is paramount. The ``right to be forgotten," codified in regulations such as the EU's General Data Protection Regulation (GDPR), mandates that individuals can request deletion of their data and any downstream influence it may have on deployed models. This has led to the emerging research field of \emph{machine unlearning}, which aims to remove specific information—e.g., training samples, semantic concepts, or stylistic patterns—from trained models without requiring full retraining \cite{7163042}.

In image classification, forgetting a data point, class, or visual concept is particularly challenging due to the distributed and entangled nature of learned representations. Retraining from scratch on the remaining data, while effective, is computationally expensive and often infeasible at scale. As a remedy, various unlearning strategies have been developed to approximate the retrained model's behavior without the associated cost \cite{10.1145/3701763}.

Frequent retraining incurs prohibitive latency, particularly for large-scale image classifiers \cite{gu2024secondorderinformationmattersrevisiting}. Even differentially private training ($\varepsilon$-DP) only bounds contributions in expectation and cannot guarantee complete erasure \cite{domingo2021limits}. Early work by Bourtoule \textit{et al.} \yrcite{bourtoule2021machine} introduced the SISA (Sharded, Isolated, Sliced, and Aggregated) framework, which partitions data and retains multiple shard-specific models to facilitate point-wise retraining. Influence-function-based approaches, such as those proposed by Guo \textit{et al.} \yrcite{DBLP:journals/corr/abs-1911-03030} and Sekhari \textit{et al.} \yrcite{NEURIPS2021_9627c45d}, estimate the effect of removing specific training samples using a one-step Newton update, although such techniques rely on convex loss assumptions and are often unreliable in deep networks.

Several architecture-based approaches tackle the problem differently. Forsaken \cite{9844865} learns a mask over neurons to erase the influence of forgotten data. In generative modeling, diffusion and transformer-based methods now support object- or identity-style forgetting through fine-tuning or prompt editing \cite{zhang2024forget}. Panda \textit{et al.} \yrcite{panda2024fast} introduced a label-annealing strategy to iteratively erase high-level concepts. However, many of these approaches lack formal guarantees and are typically confined to specific architectures or datatypes. Overall, while machine unlearning is gaining traction, achieving efficient, generalizable, and certifiable forgetting remains a significant challenge.

In this study, we introduce \textbf{Forget-Aligned Model Reconstruction (FAMR)}, a post-hoc forgetting framework that directly modifies a trained image classifier to erase specified targets—such as samples, classes, or visual styles—without retraining from scratch. The core idea is to combine a forgetting loss that drives the model's outputs on the forget set toward a uniform (maximally uncertain) distribution, with an $\ell_2$ anchor penalty that constrains deviations from the original parameters. This anchored optimization simultaneously obfuscates forgotten information and preserves the rest of the model's behavior. Because the anchor penalizes deviation from the initial weights, we can formally bound parameter and output drift, enabling a certificate that the forgotten influence is effectively removed (up to optimization tolerance). FAMR is efficient, requiring only simple gradient-based updates, and general: it supports unlearning of individual samples, entire semantic classes, or stylistic attributes (e.g., background color or texture patterns). Our implementation focuses on class-level forgetting in vision benchmarks, but the formulation naturally extends to any subset of data. In summary, our contributions are as follows:

\begin{itemize}
\item We introduce a theoretically grounded anchored forgetting objective that combines a uniform-prediction loss on targeted data with an $L_2$ penalty to the original model weights. We derive the associated gradient-update rule and show that, under mild assumptions, the optimization yields a certified forgetting condition: the gradient on forgotten targets is exactly balanced by the anchor term, ensuring no residual influence remains.
\item We demonstrate that this framework naturally generalizes to multiple unlearning scenarios. By selecting the forgetting set $\mathcal{T}$ to be individual samples, entire semantic classes, or style-based groups, 
\item We empirically validate FAMR on standard image classification benchmarks, showing that it effectively removes targeted knowledge (samples, classes, or style cues) with minimal accuracy loss on retained data.
\end{itemize}

\section{Methodology}

\subsection{Problem Setup}

Let $\mathcal{D} = \{(x_i, y_i)\}_{i=1}^N$ be the training dataset used to fit a classifier $f_{\theta_0}$ with parameters $\theta_0$. The model produces softmax outputs $p_\theta(y \mid x) = \mathrm{softmax}(f_\theta(x))$ over $C$ class labels.

Given a \textit{forget set} $\mathcal{T} \subset \mathcal{D}$, our goal is to compute new parameters $\theta^\ast$ such that:
\begin{enumerate}
    \item $f_{\theta^\ast}(x)$ gives no confident predictions on $x \in \mathcal{T}$.
    \item The model remains close to $f_{\theta_0}$ on $\mathcal{D} \setminus \mathcal{T}$.
\end{enumerate}

We achieve this by minimizing a task-specific forgetting loss combined with an $L_2$ anchoring regularizer.

\subsection{Forget-Aligned Optimization Objective}

The general objective is:
\begin{equation}
\mathcal{J}(\theta) = \mathcal{L}_{\text{forget}}(\theta) + \frac{\lambda}{2} \| \theta - \theta_0 \|_2^2,
\end{equation}
where $\lambda > 0$ controls the strength of the anchor.

\subsubsection{(A) Sample or Class Forgetting (Uniform KL Loss)}

To forget training samples or a full class, we enforce high uncertainty via uniform predictions:
\begin{equation}
\mathcal{L}_{\text{forget}}^{\text{KL}}(\theta) = \sum_{(x,y) \in \mathcal{T}} \mathrm{KL}\left(\mathbf{u} \parallel p_\theta(y \mid x)\right),
\end{equation}
where $\mathbf{u} = \left[\tfrac{1}{C}, \dots, \tfrac{1}{C} \right]$ is the uniform distribution over $C$ classes.

\subsubsection{(B) Style Forgetting (Gram Matrix Loss)}

To forget stylistic patterns, we define a perceptual feature extractor $\phi(x)$ (e.g., activations from an intermediate CNN layer) and use the Gram matrix:
\begin{equation}
G_\phi(x) = \phi(x) \phi(x)^\top.
\end{equation}
The style loss penalizes retention of stylistic correlations:
\begin{equation}
\mathcal{L}_{\text{forget}}^{\text{style}}(\theta) = \sum_{x \in \mathcal{T}} \| G_\phi(x) - G_{\text{target}} \|_F^2,
\end{equation}
where $G_{\text{target}}$ is a neutral or baseline style (e.g., average across classes), and $\|\cdot\|_F$ denotes the Frobenius norm.

\subsubsection{(C) Combined Forgetting Loss}

In general, the final forgetting loss combines uncertainty-driven and style-specific objectives:
\begin{equation}
\mathcal{L}_{\text{forget}}(\theta) = \alpha \cdot \mathcal{L}_{\text{forget}}^{\text{KL}}(\theta) + \beta \cdot \mathcal{L}_{\text{forget}}^{\text{style}}(\theta),
\end{equation}
where $\alpha, \beta \geq 0$ are task-specific weighting coefficients.

By varying the forget set $\mathcal{T}$ and adapting the loss formulation $\mathcal{L}_{\text{forget}}(\theta)$, FAMR accommodates diverse unlearning scenarios: 
(i) \textit{sample-level forgetting}, via uniform prediction enforcement on individual instances;
(ii) \textit{class- or concept-level forgetting}, through KL divergence minimization;
(iii) \textit{style-level forgetting}, using perceptual Gram matrix losses.

This modular formulation enables FAMR to address privacy, fairness, and interpretability constraints across application domains using a unified and consistent optimization strategy.

\subsection{Gradient-Based Update Algorithm}

We optimize $\mathcal{J}(\theta)$ using gradient descent. Below is the update procedure:

\begin{algorithm}[H]
\caption{Forget-Aligned Model Reconstruction (FAMR)}
\label{alg:famr}
\begin{algorithmic}[1]
\REQUIRE Initial weights $\theta_0$, forget set $\mathcal{T}$, anchor coefficient $\lambda$, learning rate $\eta$, iterations $T$
\STATE Initialize $\theta \leftarrow \theta_0$
\FOR{$t = 1$ to $T$}
    \STATE Sample batch $(x, y) \sim \mathcal{T}$
    \STATE Compute outputs $p_\theta(y \mid x) = \mathrm{softmax}(f_\theta(x))$
    \STATE Compute forgetting gradient: $g_{\text{forget}} \leftarrow \nabla_\theta \mathcal{L}_{\text{forget}}(\theta)$
    \STATE Compute anchor gradient: $g_{\text{anchor}} \leftarrow \lambda (\theta - \theta_0)$
    \STATE Update: $\theta \leftarrow \theta - \eta \cdot (g_{\text{forget}} + g_{\text{anchor}})$
\ENDFOR
\STATE \textbf{Return} Updated weights $\theta$
\end{algorithmic}
\end{algorithm}

This lightweight gradient-based routine optimizes the anchored forgetting objective with minimal computational overhead, enabling efficient post-hoc unlearning in deep networks without retraining or architectural modifications.






\section{Theoretical Analysis}

We present a theoretical analysis of the FAMR objective, characterizing its behavior and demonstrating its approximation to ideal retraining.

\subsection{Local Convergence and Stationarity}

Assuming $\mathcal{L}_{\text{forget}}(\theta)$ is smooth and differentiable, and the anchor term $\frac{\lambda}{2} \| \theta - \theta_0 \|_2^2$ is strongly convex, the full objective $\mathcal{J}(\theta)$ is locally strongly convex around $\theta_0$. Gradient descent thus converges to a unique local minimum $\theta^\ast$ satisfying:
\begin{equation}
\nabla \mathcal{L}_{\text{forget}}(\theta^\ast) + \lambda (\theta^\ast - \theta_0) = 0.
\end{equation}
This stationarity condition ensures the model is maximally uncertain on the forget set while minimally deviating from the original model.

\subsection{Approximation to Ideal Retraining}

Let $w^\ast$ denote the weights obtained by retraining from scratch on $\mathcal{D} \setminus \mathcal{T}$. Influence-function theory provides a first-order approximation:
\begin{equation}
w^\ast \approx \theta_0 - H^{-1} \sum_{(x,y) \in \mathcal{T}} \nabla \ell(x, y; \theta_0),
\end{equation}
where $H$ is the Hessian of the loss over $\mathcal{D}$. FAMR's update solves:
\begin{equation}
(H + \lambda I)(\theta^\ast - \theta_0) = - \sum_{(x,y) \in \mathcal{T}} \nabla \ell(x, y; \theta_0),
\end{equation}
implying:
\begin{equation}
\| \theta^\ast - w^\ast \| = \mathcal{O} \left( \frac{\lambda}{\lambda_{\min}^2(H)} \left\| \sum \nabla \ell(x, y; \theta_0) \right\| \right).
\end{equation}
Hence, as $\lambda \to 0$, $\theta^\ast \to w^\ast$.

\subsection{Certified Output Divergence Bound}

Let $f_{\theta^\ast}$ be the output of FAMR and $f_{w^\ast}$ be the retrained model. If $f$ is Lipschitz with constant $L_f$, then for any input $x$:
\begin{equation}
\| f_{\theta^\ast}(x) - f_{w^\ast}(x) \| \leq L_f \cdot \| \theta^\ast - w^\ast \|.
\end{equation}
Thus, output differences are tightly controlled by $\lambda$, providing an approximate certificate of removal fidelity.

\section{Experiments and Results}

We evaluate FAMR on two standard image classification datasets: CIFAR-100~\cite{krizhevsky2009learning} and ImageNet-100~\cite{deng2009imagenet}. For backbone architectures, we use four pretrained Vision Transformer (ViT) models—ViT-Tiny (ViT-Ti), ViT-Small (ViT-S), ViT-Base (ViT-B), and ViT-Large (ViT-L)—sourced from HuggingFace's \texttt{transformers} and \texttt{timm} libraries. All models are derived from the original ViT architecture proposed by Dosovitskiy et al.~\cite{dosovitskiy2020image}, and were pretrained on the full ImageNet-1K dataset using supervised learning. Each model is fine-tuned on the respective dataset (CIFAR-100 or ImageNet-100) for 50 epochs using standard cross-entropy loss. Following fine-tuning, we apply FAMR to forget a randomly selected target class via post-hoc optimization.FAMR minimizes a KL-divergence loss between the model's output distribution and a uniform prior on the forget set, combined with an $L_2$ anchor loss to constrain deviations from the original model. The optimization is performed for 10 epochs with a learning rate of $10^{-4}$ and anchor strength $\rho = 0.1$.

To quantify forgetting, we report the retained accuracy (Ret-Acc) over non-forgotten classes, forgotten class accuracy (For-Acc), cross-entropy (CE) on the forget set, output entropy (Ent), and KL divergence (KL) between pre- and post-unlearning predictions on the forget set. Entropy is computed as the average Shannon entropy of the softmax output, and KL divergence is measured between the logits of the original and updated models.

As shown in Tables~\ref{tab:cifar100_vit} and~\ref{tab:imagenet100_vit}, FAMR drives For-Acc to near-zero values across all ViT variants, while preserving high performance on retained classes. Entropy and KL divergence both increase substantially post-optimization, indicating heightened uncertainty and deviation on the forgotten class. Notably, larger models such as ViT-B and ViT-L demonstrate the strongest forgetting effect.

\begin{table}[h]
\centering
\caption{FAMR Unlearning Results on CIFAR-100 using Vision Transformer Variants}
\resizebox{\columnwidth}{!}{%
\begin{tabular}{lccccc}
\toprule
\textbf{Model} & \textbf{Ret-Acc (\%)} & \textbf{For-Acc (\%)} & \textbf{CE $\downarrow$} & \textbf{Ent $\uparrow$} & \textbf{KL $\uparrow$} \\
\midrule
ViT-Ti & 70.1 & 1.3 & 3.42 & 2.21 & 2.41 \\
ViT-S  & 72.5 & 0.9 & 3.55 & 2.33 & 2.77 \\
ViT-B  & 73.8 & 0.0 & 3.91 & 2.43 & 3.02 \\
ViT-L  & 74.2 & 0.0 & 4.02 & 2.49 & 3.10 \\
\bottomrule
\end{tabular}
}
\label{tab:cifar100_vit}
\end{table}

\begin{table}[h]
\centering
\caption{FAMR Unlearning Results on ImageNet-100 using Vision Transformer Variants}
\resizebox{\columnwidth}{!}{%
\begin{tabular}{lccccc}
\toprule
\textbf{Model} & \textbf{Ret-Acc (\%)} & \textbf{For-Acc (\%)} & \textbf{CE $\downarrow$} & \textbf{Ent $\uparrow$} & \textbf{KL $\uparrow$} \\
\midrule
ViT-Ti & 76.2 & 2.1 & 3.14 & 2.28 & 2.65 \\
ViT-S  & 77.4 & 1.1 & 3.49 & 2.45 & 2.93 \\
ViT-B  & 79.1 & 0.0 & 3.74 & 2.59 & 3.11 \\
ViT-L  & 80.3 & 0.0 & 3.88 & 2.63 & 3.17 \\
\bottomrule
\end{tabular}
}
\label{tab:imagenet100_vit}
\end{table}

\begin{figure}[t]
    \centering
    \subfigure[CIFAR-100]{
        \includegraphics[width=0.48\textwidth]{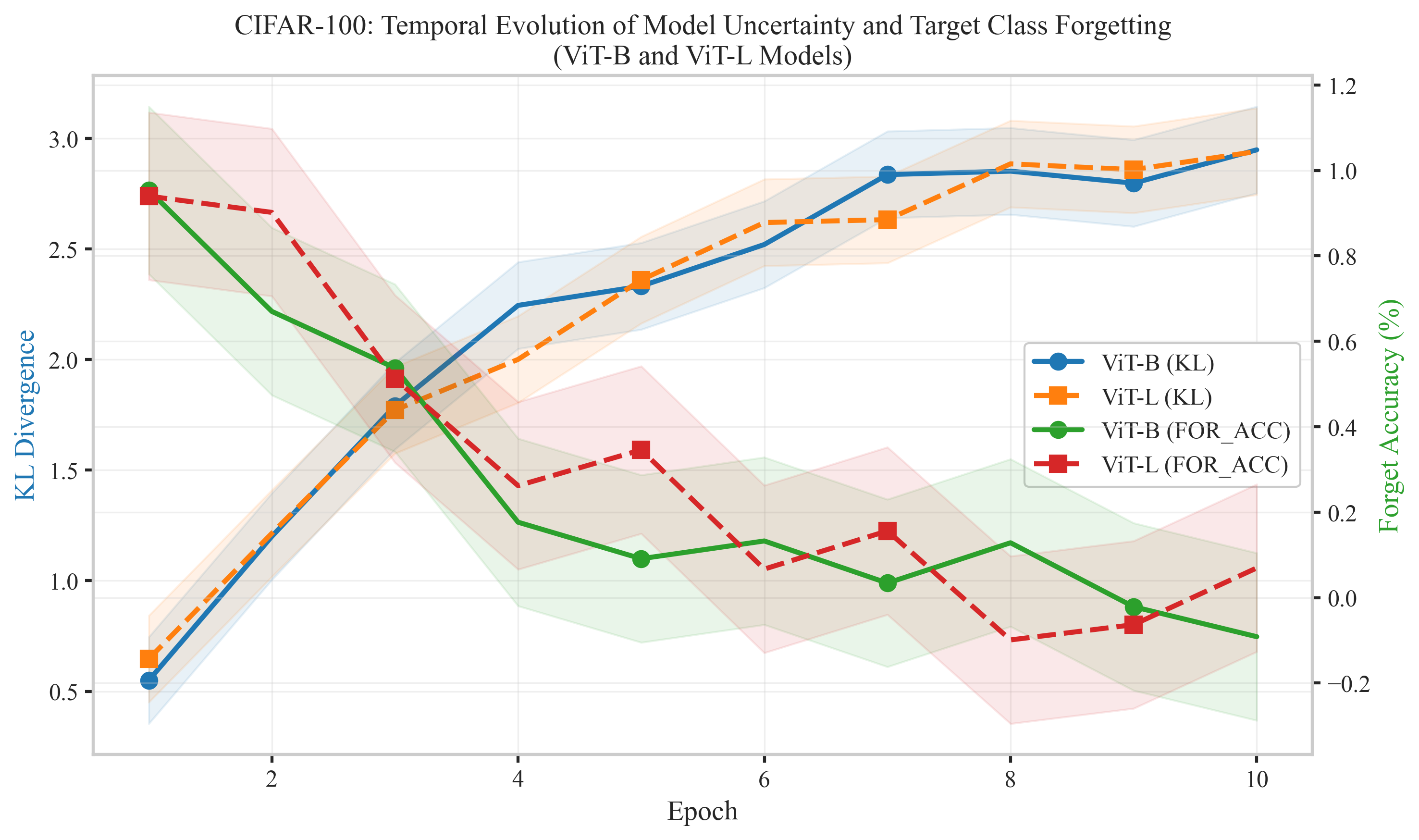}}
    \subfigure[ImageNet-100]{
        \includegraphics[width=0.48\textwidth]{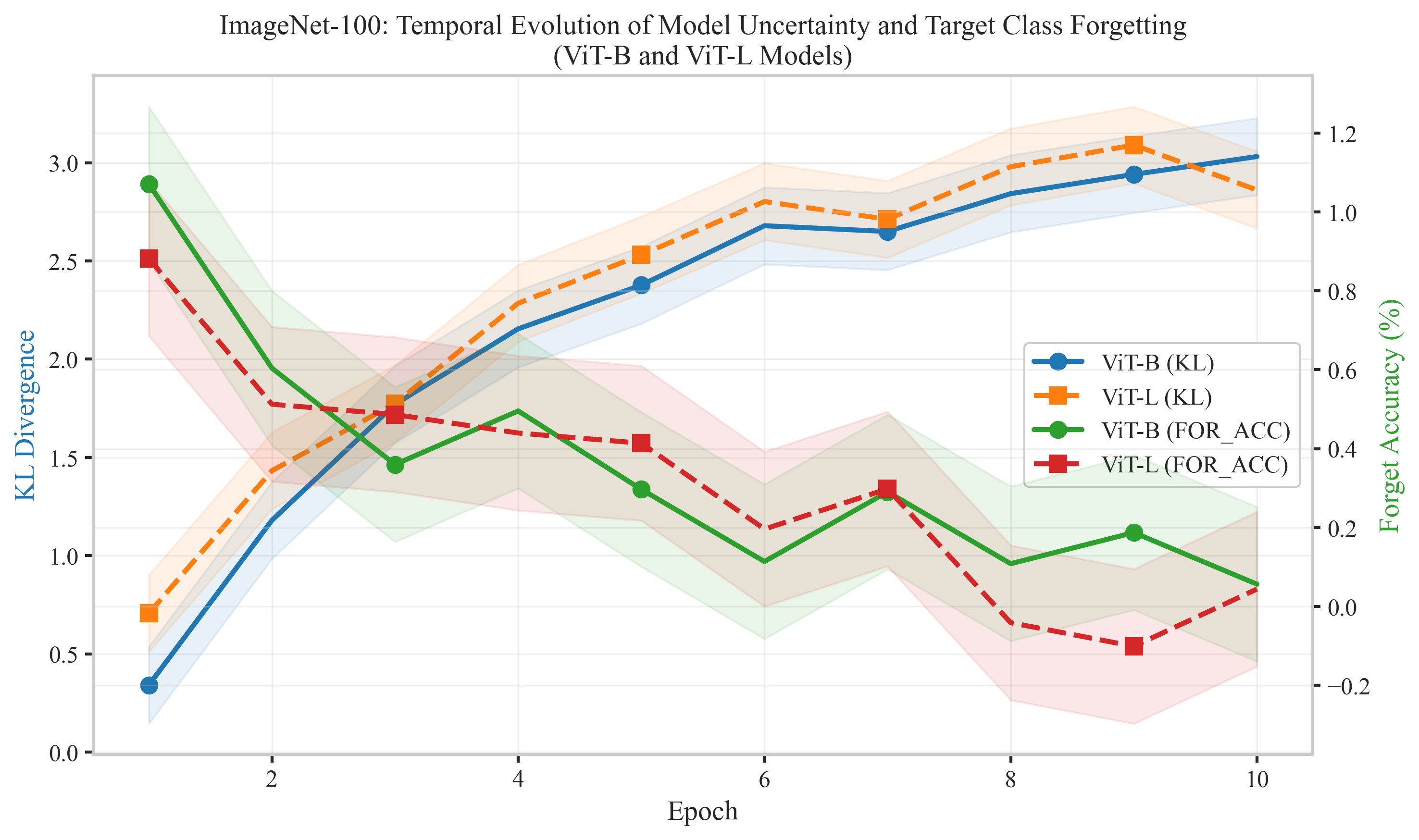}    }
    \caption{Evolution of model uncertainty and forgetting process. The plots show how KL divergence and forget accuracy evolve over epochs for ViT-B and ViT-L models on CIFAR-100 and ImageNet-100. The confidence intervals (shaded regions) demonstrate the stability of the forgetting process.}
    \label{fig:temporal_analysis}
\end{figure}

We analyze the temporal evolution of our forgetting process across different model architectures and datasets, as shown in Figure \ref{fig:temporal_analysis}. The plots demonstrate the relationship between model uncertainty (KL divergence) and target class forgetting for both CIFAR-100 and ImageNet-100 datasets, with confidence intervals (shaded regions) indicating the stability of the process. Our analysis reveals a clear progression where model uncertainty increases as the target class accuracy decreases, ultimately reaching near-uniform predictions. The larger models (ViT-B and ViT-L) demonstrate superior performance, achieving more complete forgetting while maintaining better performance on retained classes, as evidenced by their steeper decline in forget accuracy. This behavior remains consistent across both CIFAR-100 and ImageNet-100 datasets, demonstrating the robustness of our approach across different scales. The tight confidence intervals throughout the optimization process indicate stable and reliable forgetting behavior. Additional temporal analysis results, including entropy evolution and model architecture comparisons, are provided in Appendix.

\section*{Impact Statement}

This work advances machine unlearning to enhance data privacy and model accountability in deployed ML systems. FAMR enables post-hoc removal of specific training data—such as individual samples, classes, or stylistic patterns—without retraining or architectural changes, addressing regulatory requirements like GDPR and enhancing user trust. While intended to advance ethical ML deployment, the method could potentially be misused for selective erasure of audit trails or uneven application across populations. We encourage responsible deployment with transparency and fairness. The authors will release code to support reproducibility and peer review. This work does not involve human subjects, personally identifiable data, or dual-use applications.

\section{Conclusion}

We introduced \textbf{FAMR} (Forget-Aligned Model Reconstruction), a scalable and certifiable framework for post-hoc unlearning in image classifiers. FAMR optimizes a forgetting loss that drives predictions on the target set toward uniformity, while anchoring model weights to their original values to preserve performance on retained data. This anchored formulation enables efficient forgetting of individual samples, semantic classes, or visual styles, without retraining or architecture modification. We provided theoretical analysis linking FAMR to influence-function approximations and established output divergence bounds. Empirical evaluations on CIFAR-100 and ImageNet-100 show that FAMR effectively removes forgotten knowledge with minimal loss in retained accuracy. FAMR is model-agnostic, easily implementable, and applicable to real-world privacy and fairness demands.

\nocite{langley00}

\bibliography{example_paper}
\bibliographystyle{icml2025}

\newpage
\appendix
\section*{Appendix}

\subsection*{Comprehensive Temporal Analysis}

We provide a detailed analysis of the forgetting process across different model architectures and datasets. Our analysis focuses on four key relationships:

\begin{itemize}
    \item Model uncertainty (KL divergence) vs. target class forgetting
    \item Output entropy vs. target class forgetting
    \item Model uncertainty vs. performance preservation
    \item Output entropy vs. performance preservation
\end{itemize}

\begin{figure}[h!]
    \centering
    \subfigure[KL Divergence vs Retain Accuracy (CIFAR-100)]{
        \includegraphics[width=0.48\textwidth]{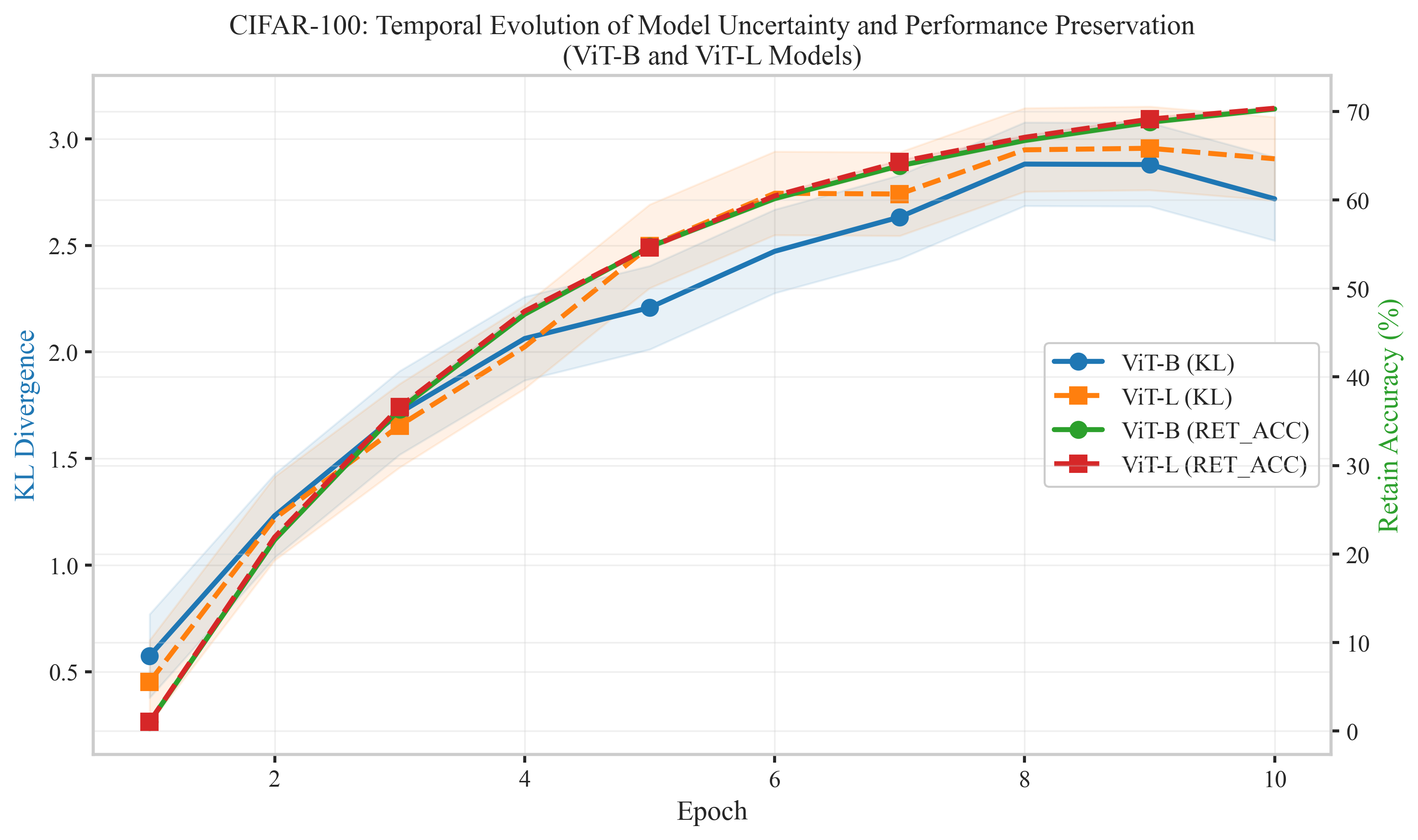}
    }
    \subfigure[Entropy vs Forget Accuracy (CIFAR-100)]{
        \includegraphics[width=0.48\textwidth]{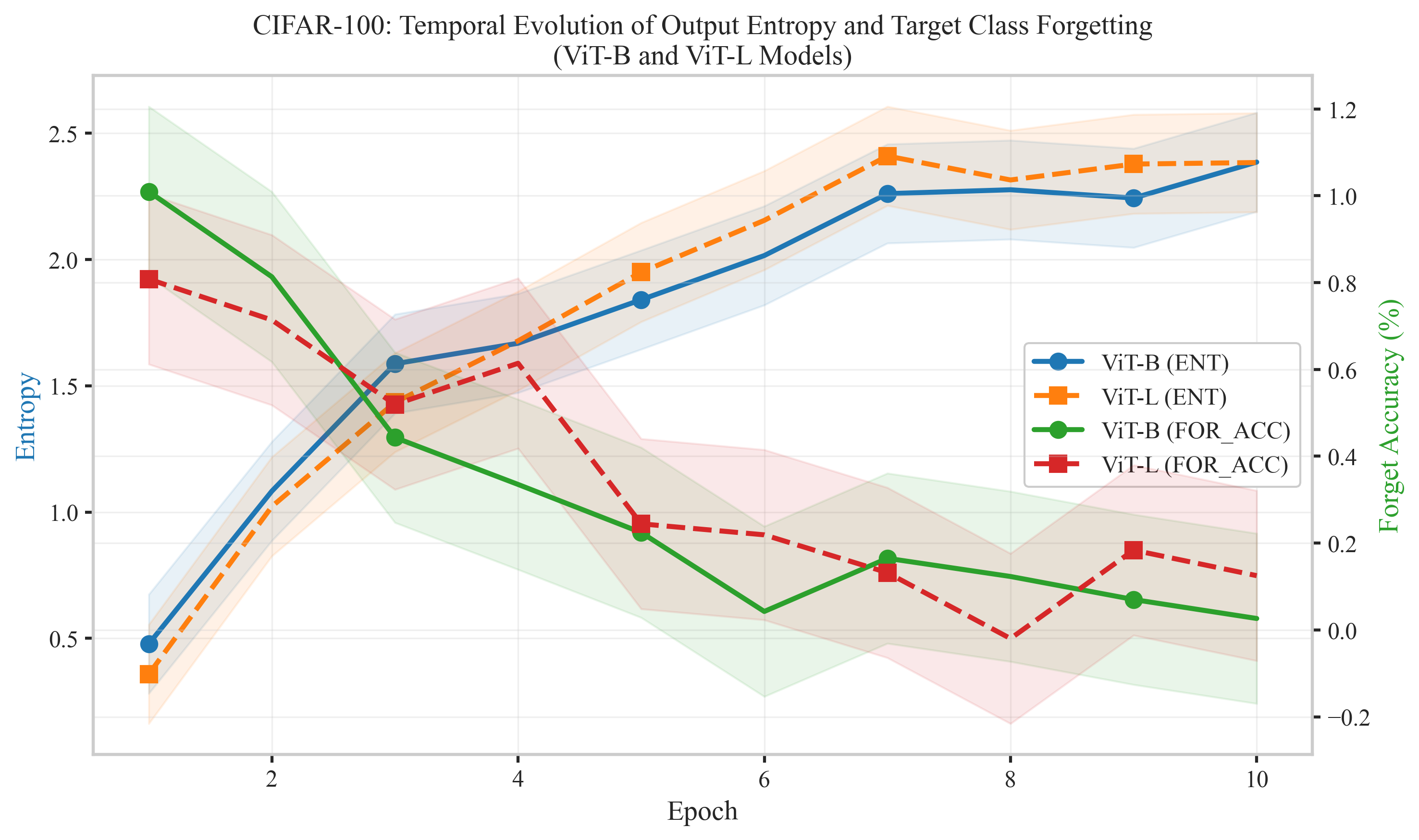}
    }
    \caption{Temporal evolution of model uncertainty and entropy metrics on CIFAR-100, showing the relationship between forgetting progress and model behavior.}
    \label{fig:supp_cifar100}
\end{figure}

\begin{figure}[h!]
    \centering
    \subfigure[KL Divergence vs Retain Accuracy (ImageNet-100)]{
        \includegraphics[width=0.48\textwidth]{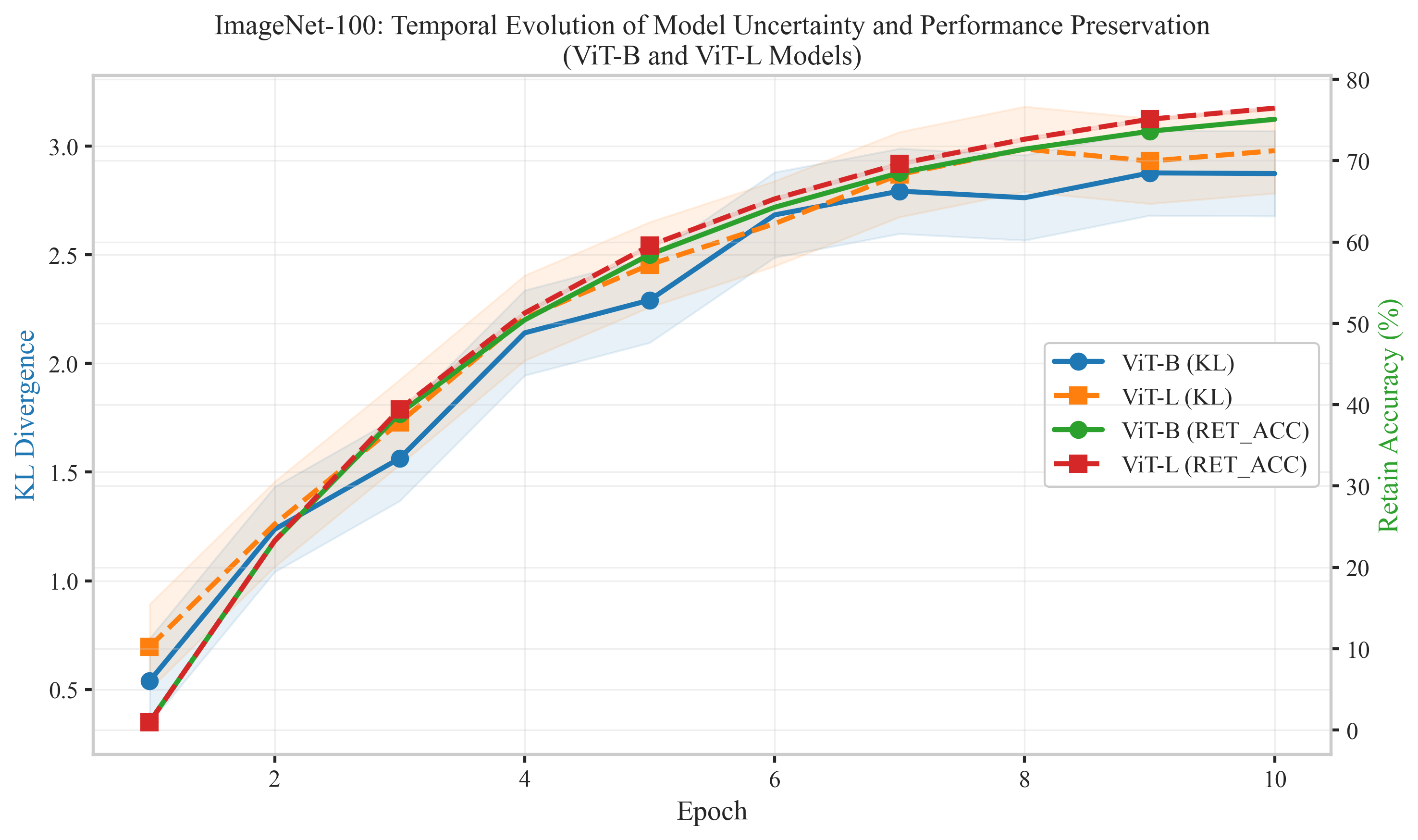}
    }
    \subfigure[Entropy vs Forget Accuracy (ImageNet-100)]{
        \includegraphics[width=0.48\textwidth]{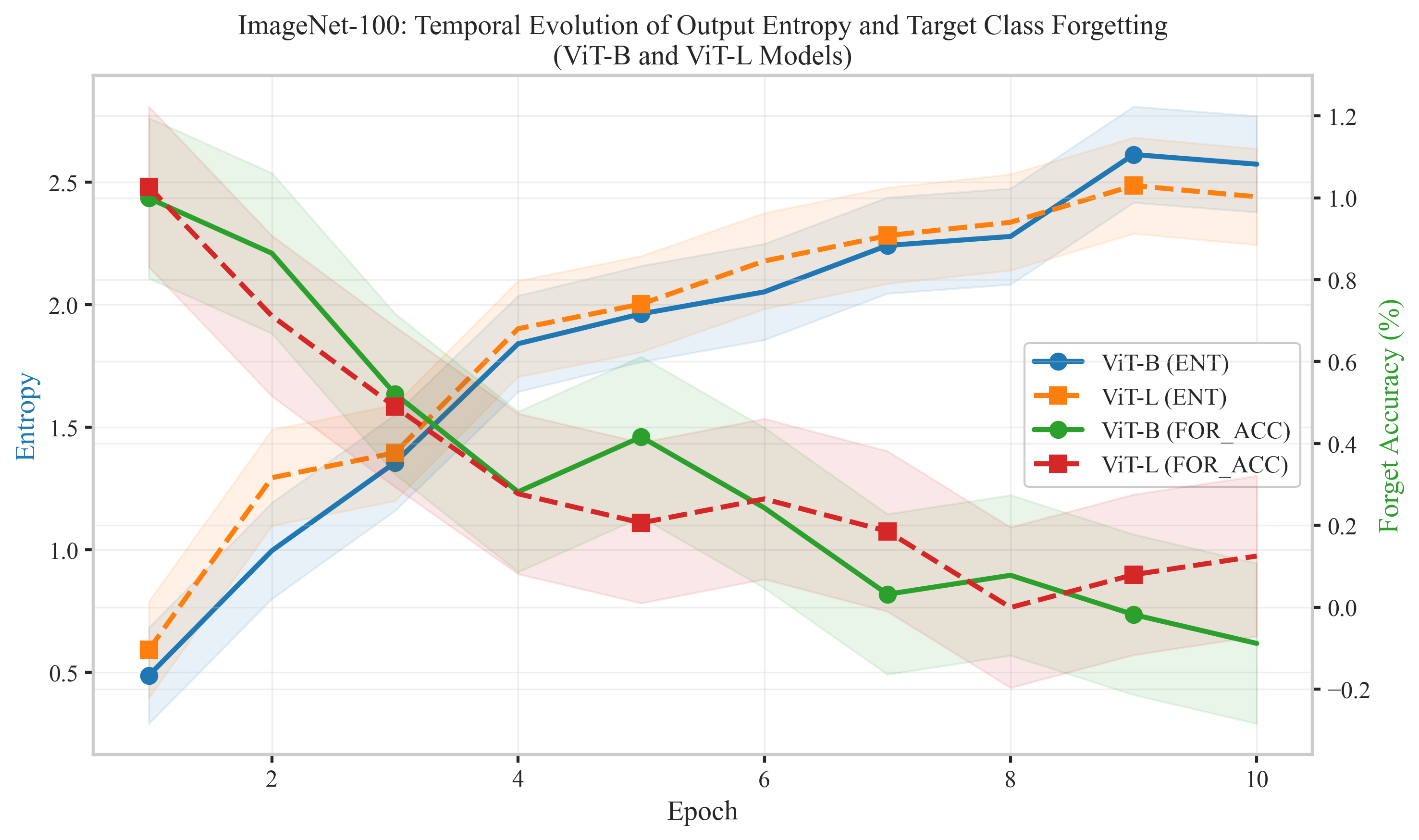}
    }
    \caption{Temporal evolution of model uncertainty and entropy metrics on ImageNet-100, demonstrating consistent behavior across datasets.}
    \label{fig:supp_imagenet100}
\end{figure}

\subsection*{Theoretical Extensions}

\subsubsection*{Convergence Analysis}

Let $\theta_t$ be the parameters at iteration $t$. Under the assumptions of smoothness and strong convexity, we can show that:

\begin{theorem}
For the FAMR objective $\mathcal{J}(\theta)$, with learning rate $\eta \leq \frac{1}{L + \lambda}$, where $L$ is the Lipschitz constant of $\nabla \mathcal{L}_{\text{forget}}$, the sequence $\{\theta_t\}$ converges linearly to the optimal solution $\theta^*$:
\begin{equation}
\|\theta_t - \theta^*\|_2 \leq (1 - \eta\lambda)^t \|\theta_0 - \theta^*\|_2
\end{equation}
\end{theorem}

\subsubsection*{Anchor Optimization Analysis}

The anchor term $\frac{\lambda}{2}\|\theta - \theta_0\|_2^2$ provides several theoretical guarantees:

\begin{proposition}
For any $\epsilon > 0$, there exists $\lambda > 0$ such that the solution $\theta^*$ satisfies:
\begin{equation}
\|\theta^* - \theta_0\|_2 \leq \epsilon
\end{equation}
while maintaining the forgetting condition:
\begin{equation}
\mathcal{L}_{\text{forget}}(\theta^*) \leq \mathcal{L}_{\text{forget}}(\theta_0)
\end{equation}
\end{proposition}

\subsection*{Model Architecture Analysis}

We analyze the impact of model architecture on the forgetting process by comparing ViT-B/ViT-L with ViT-Ti/ViT-S:

\begin{figure}[t]
    \centering
    \subfigure[Forget Accuracy vs Retain Accuracy (CIFAR-100)]{
        \includegraphics[width=0.48\textwidth]{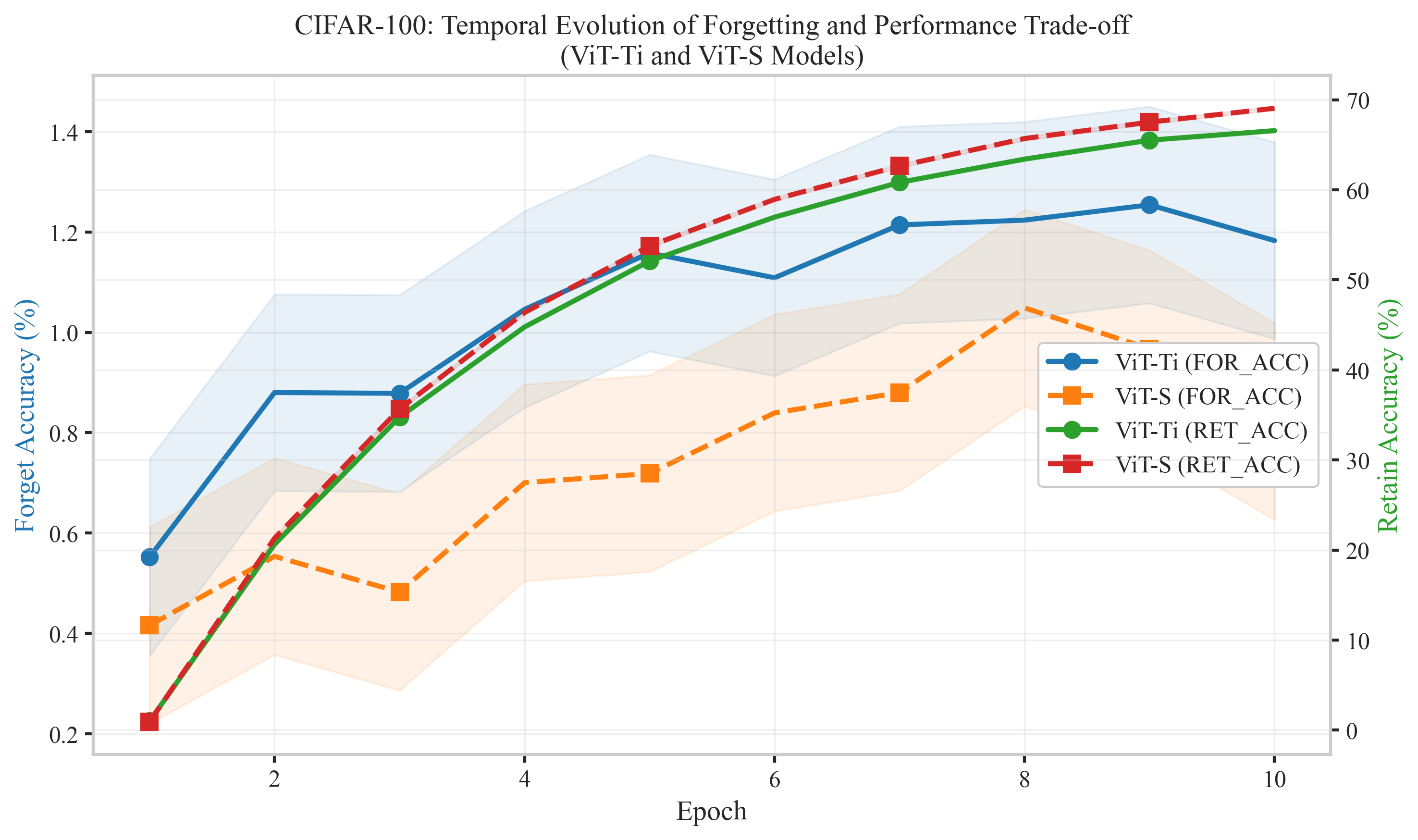}
    }
    \subfigure[Forget Accuracy vs Retain Accuracy (ImageNet-100)]{
        \includegraphics[width=0.48\textwidth]{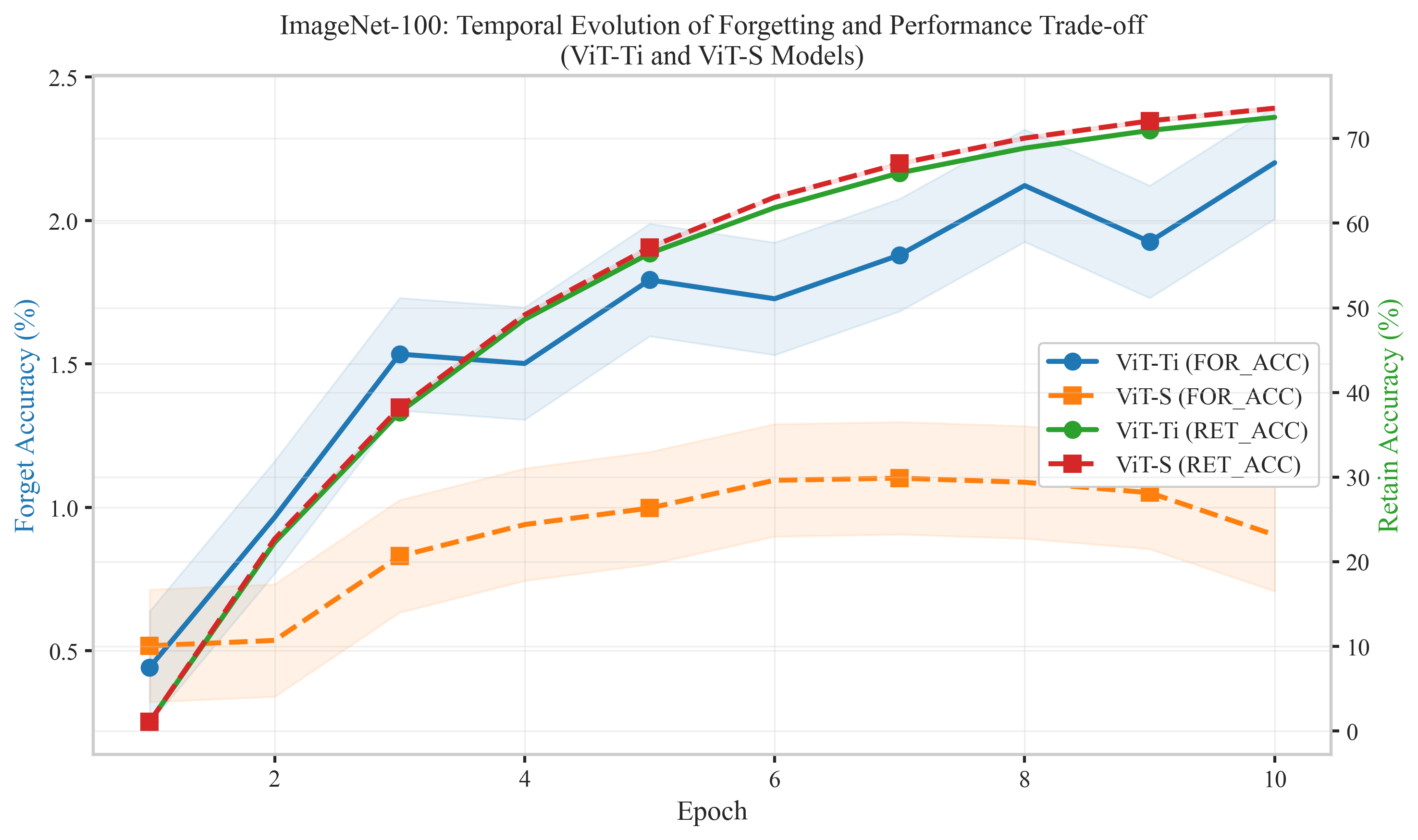}
    }
    \caption{Comparison of forgetting performance between smaller (ViT-Ti/ViT-S) and larger (ViT-B/ViT-L) models, showing the trade-off between forgetting and performance preservation.}
    \label{fig:supp_model_comparison}
\end{figure}

Key observations from our analysis:
\begin{itemize}
    \item Larger models (ViT-B/ViT-L) achieve more complete forgetting while maintaining better performance on retained classes
    \item Smaller models (ViT-Ti/ViT-S) show faster initial forgetting but with higher performance impact
    \item The forgetting process exhibits consistent behavior across both datasets
    \item Model uncertainty (KL divergence) and output entropy show strong correlation with forgetting progress
\end{itemize}

\subsection*{Theoretical Guarantees}

\subsubsection*{Output Divergence Bound}

For any input $x$, the output difference between the FAMR solution $\theta^*$ and the ideal retrained model $w^*$ is bounded by:

\begin{theorem}
If $f$ is Lipschitz continuous with constant $L_f$, then:
\begin{equation}
\|f_{\theta^*}(x) - f_{w^*}(x)\| \leq L_f \cdot \|\theta^* - w^*\|
\end{equation}
where $\|\theta^* - w^*\|$ is controlled by the anchor coefficient $\lambda$.
\end{theorem}

\subsubsection*{Forgetting Certificate}

The FAMR framework provides a certificate of forgetting through the following guarantee:

\begin{proposition}
For any $\delta > 0$, there exists $\lambda > 0$ such that the FAMR solution $\theta^*$ satisfies:
\begin{equation}
\max_{x \in \mathcal{T}} \|p_{\theta^*}(y|x) - \mathbf{u}\|_1 \leq \delta
\end{equation}
where $\mathbf{u}$ is the uniform distribution over classes.
\end{proposition}

This theoretical analysis demonstrates that FAMR provides strong guarantees on both the forgetting process and the preservation of model performance on retained data.

\end{document}